\documentclass[letterpaper, 10 pt, conference]{ieeeconf}  

\IEEEoverridecommandlockouts                              
\usepackage{svg}
\usepackage{amsmath}
\usepackage{graphicx}
\usepackage{url}
\graphicspath{{figures/}}
\usepackage{soul} 

\title{\LARGE \bf
Incorporating End-to-End Speech Recognition Models for Sentiment Analysis
}

\author{Egor Lakomkin$^{1}$, Mohammad Ali Zamani$^{1}$, Cornelius Weber$^{1}$, Sven Magg$^{1}$ and Stefan Wermter$^{1}$
\thanks{$^{1}$Knowledge Technology, Department of Informatics, \newline University of Hamburg, Vogt-Koelln-Str. 30, 22527  Hamburg, Germany.\newline
http://www.informatik.uni-hamburg.de/WTM/
\newline
        {\tt\small $\{$lakomkin, zamani, weber, magg, wermter$\}$ @informatik.uni-hamburg.de }}%
}

\begin{document}

\maketitle
\thispagestyle{empty}
\pagestyle{empty}

\begin{abstract}

 Previous work on emotion recognition demonstrated a synergistic effect of combining several modalities such as auditory, visual, and transcribed text to estimate the affective state of a speaker. Among these, the linguistic modality is crucial for the evaluation of an expressed emotion. However, manually transcribed spoken text cannot be given as input to a system practically. We argue that using ground-truth transcriptions during training and evaluation phases leads to a significant discrepancy in performance compared to real-world conditions, as the spoken text has to be recognized on the fly and can contain speech recognition mistakes. In this paper, we propose a method of integrating an automatic speech recognition (ASR) output with a character-level recurrent neural network for sentiment recognition. In addition, we conduct several experiments investigating sentiment recognition for human-robot interaction in a noise-realistic scenario which is challenging for the ASR systems. We quantify the improvement compared to using only the acoustic modality in sentiment recognition. We demonstrate the effectiveness of this approach on the Multimodal Corpus of Sentiment Intensity (MOSI) by achieving 73,6\% accuracy in a binary sentiment classification task, exceeding previously reported results that use only acoustic input. In addition, we set a new state-of-the-art performance on the MOSI dataset (80.4\% accuracy, 2\% absolute improvement).

\end{abstract}

\section{INTRODUCTION}
 \par Speech emotion and affect recognition are crucial aspects for a coherent human-robot interaction and have recently received growing interest in the research community \cite{Zadeh2017TensorAnalysis}. The quality of human-robot interaction could be improved significantly if a robot was able to consistently evaluate the emotional state of a person and its dynamics. For instance, if a robot was able to detect that a person is speaking in an angry way, it could use this information as a sign to adjust its behavior \cite{Lakomkin2018EmoRL:Learning}.
 
 \par Humans integrate information from several input modalities, such as acoustic and visual, to estimate the emotional state of the speaker \cite{Picard1997AffectiveComputing}. Recent computational models fuse different sources of information to yield  better and more robust results. 
 For example, combining visual, linguistic and acoustic modalities resulted in state-of-the-art performance on sentiment and emotion recognition tasks and it can be observed that the linguistic modality has the biggest contribution in the overall blend \cite{Zadeh2017TensorAnalysis}. However, most experiments assume that the ground-truth (manually transcribed) spoken text transcriptions are available during training and testing phases. We argue that this setup differs from the real-life condition, as we usually do not have access to the transcribed text in real time. In practice in the best case, we only extract an approximation of it: multiple hypotheses of the speech recognition system. Also, training on clean texts might result in overstating of the model's performance and degradation during testing when we use even state-of-the-art speech recognition models. 
 
  \begin{figure}[t]
  \includegraphics[width=\linewidth]{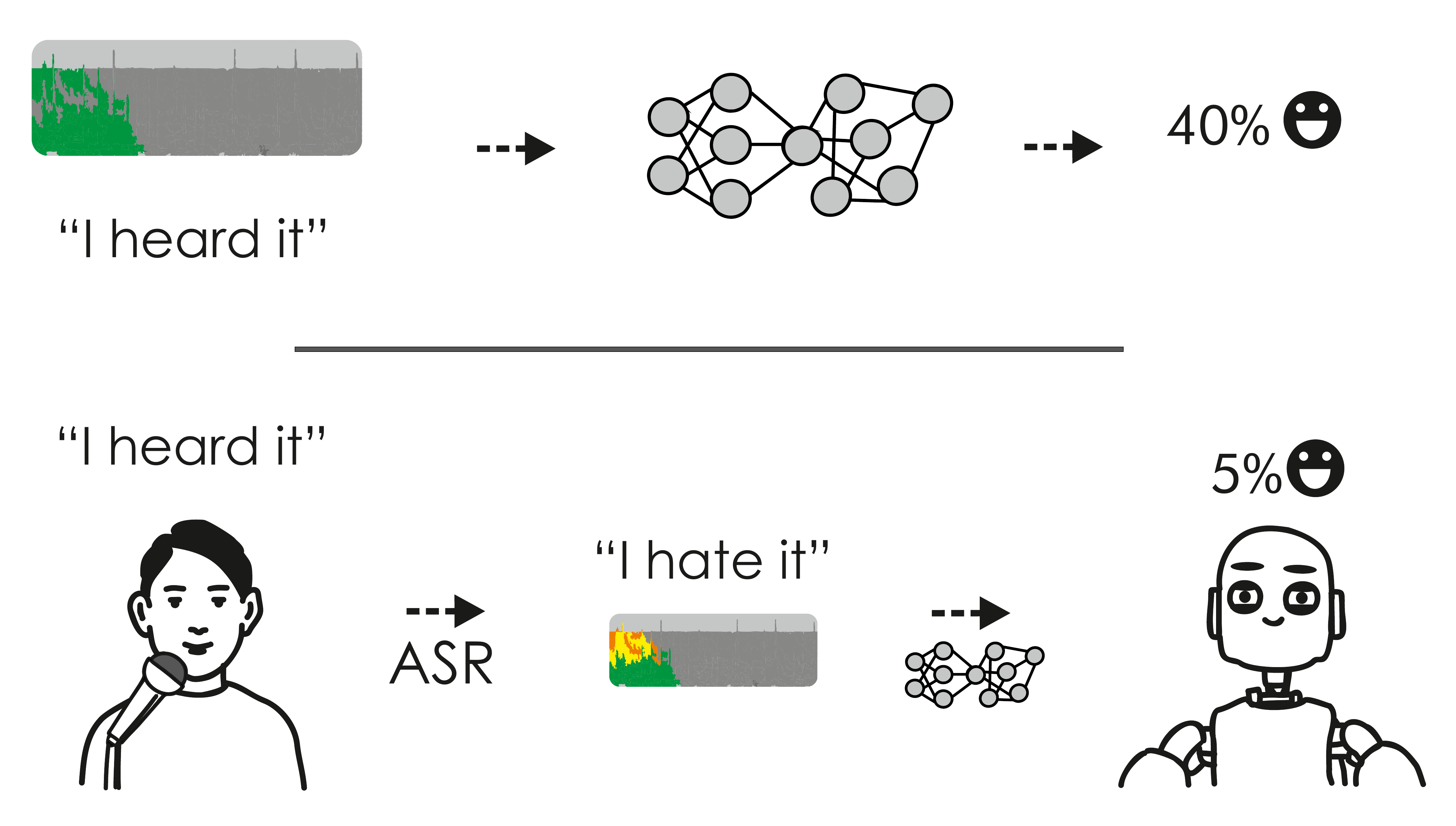}
  \caption{ Illustration of the core issue that we address in the paper. Sentiment recognition models are usually trained and evaluated using the manually transcribed text. In practice, we have to use automatic speech recognition systems to extract spoken text, which can contain errors affecting the overall performance. }
  \label{fig:system_main}
\end{figure}
 
 %
 
 \par This paper is built on our previous work \cite{Lakomkin2018EmoRL:Learning,Lakomkin2018OnNetworks} on the application of neural emotion recognition models in the context of human-robot interaction. In this work, we conduct several experiments to address the discrepancy in the performance of a spoken sentiment recognition when the manually transcribed spoken text is not available. Our contribution is two-fold: a) We evaluate the performance of neural joint acoustic-linguistic sentiment recognition models in a human-robot interaction setup when the spoken text transcriptions are not available; b) We propose and evaluate the incorporation of recurrent character-level language model representations of spoken text for sentiment modeling to adapt to situations when the ASR output might produce word outputs with mistakes due to noise. We also analyze the models' performance in acoustically clean conditions and when re-recorded on a robotic head, for the Multimodal Corpus of Sentiment Intensity (MOSI) dataset for sentiment identification. We compare three setups for spoken text extraction: 1) training our own end-to-end character-level neural speech recognition system, 2) using Google ASR API, 3) using ground-truth transcriptions, which we consider as the upper bound in our experiments.

\par The paper is organized as follows: section II introduces related work and section III describes our neural sentiment recognition model. Section IV outlines the methodology including the description of our neural ASR and the data used to train it. Section V introduces the conducted experiments on the original MOSI data and on the robot head recorded data in noise-realistic conditions.

\section{RELATED WORK}
\par Integration of multiple modalities like vision, auditory and linguistic with deep neural networks significantly boosted the overall performance of sentiment and emotion recognition \cite{Rozgic2012EnsembleRecognition}. For instance, individual modalities' representations and their paired combinations were fused by an outer tensor product \cite{Zadeh2017TensorAnalysis}.  The multi-attention recurrent network employs an attention method to model the integration of different modalities as has been done in multiple other works \cite{Zadeh2018Multi-attentionComprehension}. Conditioning on the context was shown beneficial  on emotion and sentiment recognition \cite{Poria2017Context-DependentVideos}. As some modalities can have different contributions a certain time step, a gating mechanism which is trained with reinforcement learning, is introduced to switch on or off a particular modality's input \cite{Chen2017MultimodalLearning}. 

\par In this work, we are focusing on the acoustic and linguistic modalities, considering situations when the speaker might not be directly observable. Jin et al. \cite{Jin2015SpeechFeatures} used various hand-crafted acoustic and lexical features followed by late decision fusion for classification. Multimodal word-level alignment produced state-of-the-art results on the emotion and sentiment recognition tasks \cite{Gu2018MultimodalAlignmentb}. Automatic generation of ensemble trees with SVM classifiers as nodes was applied for affective analysis \cite{Rozgic2012EnsembleRecognition}. Hybrid attention mechanism was introduced  to fuse acoustic and linguistic information \cite{Gu2018HybridClassification}. Aldeneh et al. \cite{Aldeneh2017PoolingValenceb} evaluated several end-to-end approaches of pooling lexical and acoustic features extracted by recurrent neural networks, which encode each modality separately for speech valence estimation. Attention-based convolutional neural networks were proposed for acoustic-only emotion recognition \cite{Neumann2017AttentiveSpeech}. Etienne et al. \cite{Etienne2018SpeechAdjustment} achieved state-of-the-art results among systems using only audio modality combining convolutional and recurrent layers. However, Schuller et al. \cite{Schuller2018TheBeats} compared off-the-shelf acoustic feature extractors with end-to-end approaches and demonstrated that end-to-end methods still do not consistently surpass the handcrafted representations on the paralinguistic tasks. Several ways of transfer learning to encode acoustic signals were proposed recently: tuning audio representations trained initially for other auxiliary tasks, like gender and speaker identification \cite{Gideon2017ProgressiveRecognition} or speech recognition \cite{Fayek2016OnRecognition, Lakomkin2017ReusingRecognition}.

\par The main difference between our work and the previous research is that we do not assume that the ground-truth transcriptions are given as input to the model. We observe that the linguistic modality makes the biggest contribution  to the overall classification \cite{Blanchard2018GettingModalities}. In this work, we assume to have access only to the acoustic input, and spoken text is therefore extracted by a separate module. 

\section{SENTIMENT RECOGNITION MODEL}

\begin{figure}[t]
  \includegraphics[width=\linewidth]{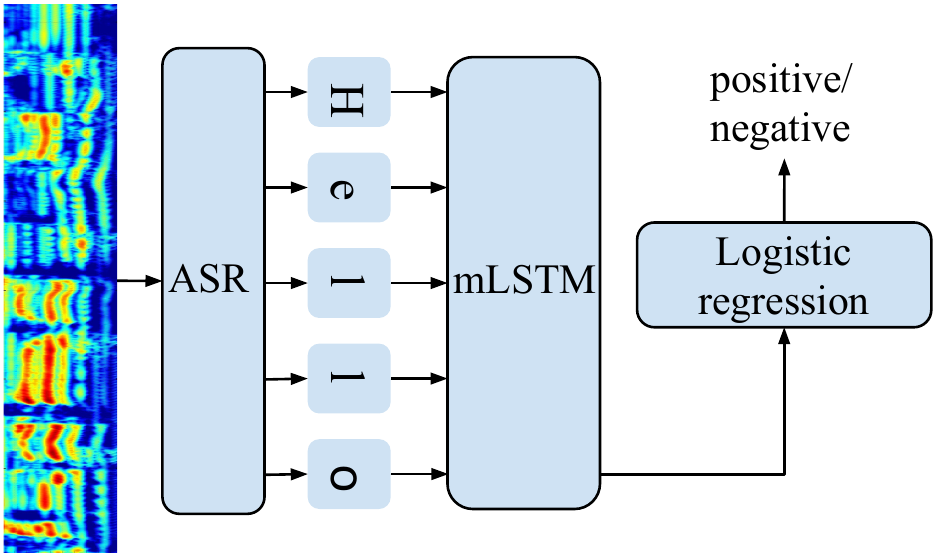}
  \caption{Our sentiment recognition model based on the ASR character output. The multiplicative LSTM model is used to encode spoken text to a fixed-length vector with a logistic regression on top modeling the sentiment. The mLSTM model is pretrained in an unsupervised way on the Amazon reviews with a language-modelling objective.}
  \label{fig:overallmodel}
\end{figure}
\par This section describes our sentiment recognition model. Firstly, we outline the neural spoken text representation network, followed by our proposed methods to extract transcriptions. We conclude with acoustic features description, which we used for joint acoustic-linguistic sentiment analysis.

\subsection{Model Architecture}
\par The essential part of our architecture is the character-level recurrent neural network for spoken text encoding. We use a single layer multiplicative LSTM \cite{Krause2016MultiplicativeModelling} (mLSTM) model with 4,096 nodes and we use the hidden state corresponding to the last input character to represent the whole textual input. The mLSTM model is trained on the vast amounts of Amazon reviews in a language modelling setup \cite{Radford2017LearningSentiment}. This model with a linear classifier on top showed state-of-the-art performance on the Stanford Sentiment Treebank outperforming more complex architectures. This indicates that even though the mLSTM model was trained fully unsupervised, it was capable to capture the concept of sentiment just by learning to predict the next character given the context. Our main intuition to use this character-level model for spoken sentiment recognition opposes to the majority of previous work incorporating word-level processing is two-fold: a) a character-level model is capable of dealing with the spelling mistakes or out-of-vocabulary words produced by the ASR model, b) the representations learnt by the mLSTM on the reviews could be useful as well in the spoken sentiment analysis.
\par 
The overall architecture is shown in Figure \ref{fig:overallmodel}. We feed the input text representation to an L2-regularized logistic regression for binary sentiment classification. The same architecture is used to train the acoustic model, where we use off-the-shelf acoustic feature descriptors (see section III.C) instead of the pre-trained mLSTM as a feature extractor. In addition, we fuse sentiment predictions of acoustic and linguistic classifiers by computing their weighted combination. We tune hyperparameters (logistic regression regularization strength and classifiers fusion weight) on the validation data.

\subsection{Spoken Text Extraction}

\par In our experiments, we train our own end-to-end ASR model (see section IV). Our ASR model computes character probabilities for each timestep and the final transcription is extracted by simply taking the most probable character for each frame (greedy decoding). The only post-processing steps we perform are a) we merge together blank symbols (used by the ASR model to denote a non-speech character or a change between different characters and displayed here as '\_') and character repetitions (\textit{aaa\_\_bbb - a\_b}), b) capitalized characters are lowercased with a space imputed in front  (\textit{H\_i\_H\_o\_w\_A\_r\_e\_Y\_o\_u - hi how  are you}). We do not apply any language model to correct the potential spelling mistakes of the model. For comparison, we extract the most probable transcription using Google Web Speech API\footnote{\url{https://pypi.org/project/SpeechRecognition/}}.

\subsection{Acoustic Feature Extraction}

We use the \textit{COMPARE 2016} feature set extracted by the OpenSMILE toolkit \cite{Eyben2013RecentExtractor}.
This feature set contains 6,373 features resulting
from the computation of various functionals (for example mean, standard deviation, maximum value) over low-level
descriptors (like mel cepstral coefficients, pitch, loudness, etc.) described in \cite{Weninger2013OnCommon.}.

\section{SPEECH RECOGNITION MODEL}
\begin{figure}
  \vspace{10pt}
  \includegraphics[width=\linewidth]{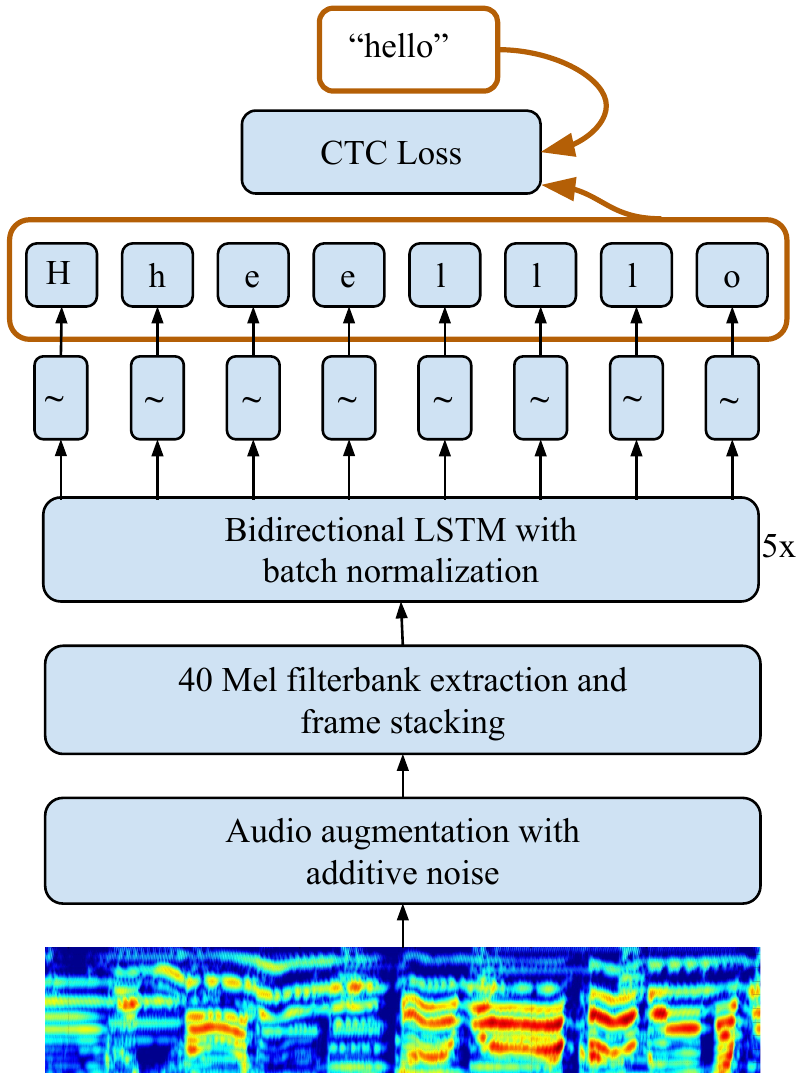}
  \caption{Architecture of the ASR model used in this work. The model is trained end-to-end by mapping mel-spectrograms to characters using a stack of LSTM layers and CTC loss function. }
  \label{fig:asrmodel}
\end{figure}
To extract spoken text from the acoustic signal we train an end-to-end neural automatic speech recognition system. In this section, we outline the ASR architecture, describe the data preprocessing and feature extraction pipeline, and datasets used for training.

\subsection{Architecture}
Our ASR model (see Figure \ref{fig:asrmodel}) is based on several stacked Long Short Term Memory (LSTM) \cite{Hochreiter1997LongMemory} recurrent layers \cite{Amodei2016DeepMandarin}. The model contains five bi-directional LSTMs with batch normalization layers in between \cite{Amodei2016DeepMandarin} processing log mel-spectrograms extracted from input audio. We use 40 mel coefficients, extracted using a Hamming window of 25ms width and 10ms stride. We stack three consecutive speech frames resulting in 120 features for each timestep. Frame stacking greatly speeds up the training and makes it more stable as the input and output are three times shorter. Recurrent layers are followed by a fully connected layer with a softmax activation on top, modelling the character probability distribution for each speech frame. Along with the standard 26 English characters, we introduce capital characters to the overall set denoting the beginning of words for the model. Overall, our model has around 61 million parameters. Connectionist Temporal Classification (CTC)  \cite{Graves2006ConnectionistClassification} is used as a loss criterion to measure how the alignment produced by the network matches to the ground-truth transcription. 

\par The Stochastic Gradient Descent optimizer is used in all experiments with a learning rate of 0.0001 and a Nesterov momentum value of 0.9, clipping the norm of the gradient at the level of 400 with a batch size of 24. During the training, we apply learning rate annealing with a factor of 1.1.  We apply the SortaGrad algorithm \cite{Amodei2016DeepMandarin} during the first epoch by sorting utterances by their duration \cite{Hannun2014DeepRecognition}. We select the model with the best word error rate measured on the LibriSpeech validation set (combined clean and noisy splits) to prevent model overfitting. We train the model on two GTX1080TI, and it takes around five days to train the model until convergence.

\subsection{Data Augmentation}

Previous research in end-to-end speech recognition demonstrated the importance of introducing random perturbations into the speech signal like a change of pitch, tempo, loudness, and adding noise \cite{Lakomkin2018OnNetworks, Amodei2016DeepMandarin, Hannun2014DeepRecognition, Zhou2017ImprovedRecognition}. Since such perturbations do not alter the target label (spoken text in the case of speech recognition, or an emotion category), they can be conveniently applied with some occurrence probability during training. Data augmentation can be considered also as a way to increase the training data size. In the case of human-robot interaction, it is crucial to have a noise-robust model due to the presence of a robot's ego-noise or background noise.

\par In our experiments, we 1) change the tempo of the recording by sampling the speed factor uniformly in a range of [85, 120] percent, 2) change the loudness of the recording by sampling gain uniformly in a range of [-6, 5] dB, 3) add random background noise, where non-speech noise samples are selected from Google's AudioSet\footnote{\url{https://research.google.com/audioset/}}, by sampling the noise-to-signal ratio uniformly in the range [0.1, 0.4] and mixing it with the original utterances resulting in over 530,000 samples of 10 seconds in length, and 4) perturb vocal tract length in the range [0.9, 1.1].

\subsection{Speech Data}
In our experiments, we use only freely available datasets.
We concatenate five datasets to train the ASR model: LibriSpeech, TED-LIUM v3, Mozilla Common Voice, Google Speech Commands v2 and VoxForge. LibriSpeech \cite{Panayotov2015Librispeech:Booksb} contains around 1,000 hours of English-read speech from audiobooks. TED-LIUM v3 \cite{Hernandez2018TED-LIUMAdaptation} is a dataset composed of transcribed TED talks, containing 452 hours of speech and 2,351 speakers. VoxForge is an open-source collection of transcribed recordings collected using crowd-sourcing. We downloaded all English recordings\footnote{\url{http://www.repository.voxforge1.org/downloads/SpeechCorpus/Trunk/Audio/Original/48kHz_16bit/}}, which are around 100 hours of speech. Common Voice\footnote{\url{https://voice.mozilla.org/}} is a crowdsourced dataset, where utterances were collected through a web interface. Each participant was asked to pronounce a predefined text and submit it to the website. In addition, other volunteers were asked to check if the spoken text matches the actual requested one. Overall, Common Voice contains around 300 hours of validated speech data.  Google Speech Commands contains 100,000 short recordings with only one word pronounced (out of 30 possible ones) by a variety of speakers.
Overall, 850,000 utterances containing ~1,600 hours of speech from more than 3,000 speakers are used to train the ASR model.  We conduct no preprocessing other than the conversion of recordings to WAV format with single-channel 16-bit signed integer format and a sampling rate of 16,000. Utterances longer than 15 seconds are filtered out due to GPU memory constraints.

\section{Experiments and analysis}

\begin{figure}[t]
  \includegraphics[width=\linewidth]{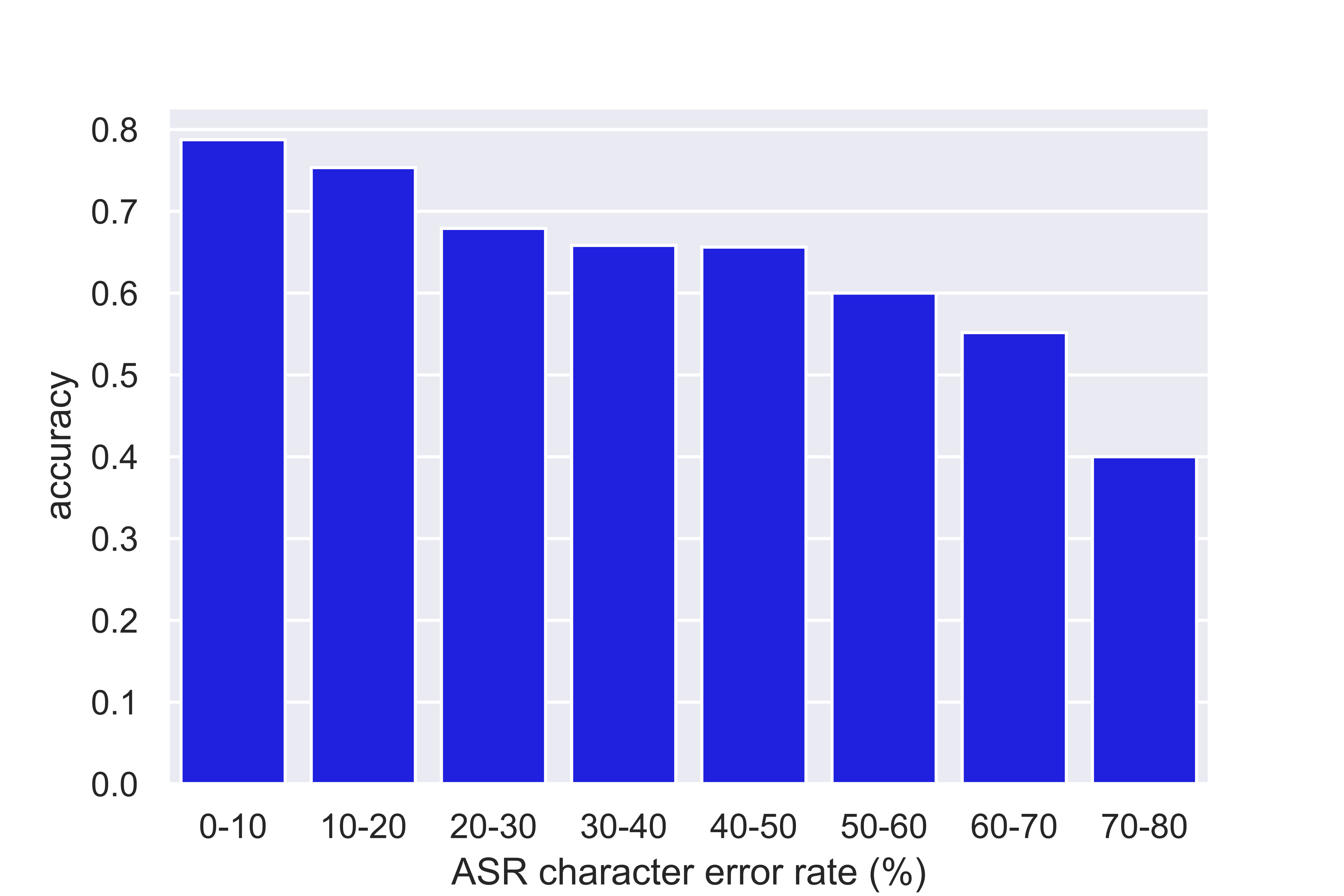}
  \caption{Histogram plot visualizing the dependency between the character error rate of our trained ASR system and the accuracy of the sentiment recognition model. We note that even for samples with 20-30\% character error rate the accuracy score is greater than 70\%.  }
  \vspace{-5mm}
  \label{fig:cer_accuracy}
\end{figure}

\begin{table*}[tbp]
\centering
\caption{Sentiment prediction results on the CMU-MOSI test set. The best result of the model which does not use ground-truth transcriptions is highlighted in bold. Text source denotes how the spoken text was extracted: either ground-truth from the MOSI data or ASR output (Google Web Speech API or our ASR) was used. MOSI-Soundman is the MOSI dataset re-recorded in our lab emulating human-robot interaction scenario. }
\label{table:results}
\begin{tabular}{cccccc}
Model     									  & Dataset  & Modalities & Text Source  & Accuracy & F-Score    \\ \hline

Tensor fusion network, Zadeh et al. \cite{Zadeh2017TensorAnalysis} & MOSI & audio + text & ground-truth & 74.6\% &  74.5\% \\
MARN, Zadeh et al. \cite{Zadeh2018Multi-attentionComprehension} & MOSI & text & ground-truth & 77.1\% & 77.0\% \\ 
Word-level alignment, Gu et al. \cite{Gu2018MultimodalAlignmentb} & MOSI & text+audio+vision & ground-truth & 76.4\% &  76.8\% \\
Recurrent multi-stage fusion, Liang et al. \cite{Liang2018MultimodalFusion} & MOSI & text+audio+vision & ground-truth & 78.4\% &  78.0\% \\
Ours char-RNN + LogReg & MOSI & text & ground-truth & 80.4\% & 79.8\% \\
\hline
Ours, char-RNN + LogReg & MOSI & audio &  - & 54.8\% & 54.1\% \\ 
Ours, char-RNN + LogReg & MOSI-Soundman &  audio & - & 53.6\% & 53.4\% \\ \hline
Ours, char-RNN + LogReg & MOSI & text &  our ASR & 69.9\% & 68.7\% \\ 
Ours, char-RNN + LogReg & MOSI & text & Google ASR & 69.6\% & 69.3\% \\
Ours, char-RNN + LogReg (2x models fused) & MOSI & text & Google ASR + our ASR & 72.3\% & 71.9\% \\ 
Ours, char-RNN + IS16 + LogReg (3x models fused) & MOSI & text + audio &  Google ASR + our ASR & \textbf{73.6\%} & \textbf{73.1\%} \\
\hline
Ours, char-RNN + LogReg & MOSI-Soundman & text & our ASR &  58.4\% & 58.2\% \\ 
Ours, char-RNN + LogReg & MOSI-Soundman &  text & Google ASR & 67.9\% & 67.7\% \\ 
Ours, char-RNN + IS16 + LogReg (2x models fused) & MOSI-Soundman & text + audio & Google ASR  & 70.1\% & 69.7\% \\ 
Ours, char-RNN + IS16 + LogReg (3x models fused) & MOSI-Soundman & text + audio & Google ASR + our ASR  & 70.2\% & 69.8\% \\

\hline

\end{tabular}
\end{table*}

In this section, we outline the data used in our experiments and the evaluation protocol and metrics, followed by the evaluation results and comparisons to previous work.

\subsection{Data and Evaluation Measure}

\subsubsection{CMU-MOSI}
 This dataset is a multimodal sentiment intensity and subjectivity dataset consisting of
93 review videos in English with 2,199 utterance segments collected from YouTube \cite{Zadeh2016MOSI:Videosb}. Each segment is labelled
by five individual annotators between -3 (strong negative) and +3 (strong positive). We binarize the labels (positive and negative) based on the sign of
the annotations' average to compare to the previously published methods. We use an 80\%-20\% training-testing speaker-independent split following the same strategy as in previous work \cite{Zadeh2017TensorAnalysis}, leaving 10\% of training data for validation.
Specifically, there are 1,279 utterances for training, 233 for validation and 686 utterances for testing. We report accuracy and macro F1-score calculated over the test set.

\subsection{Human Robot Simulation}
\begin{figure}[tbp]
  \includegraphics[width=\linewidth]{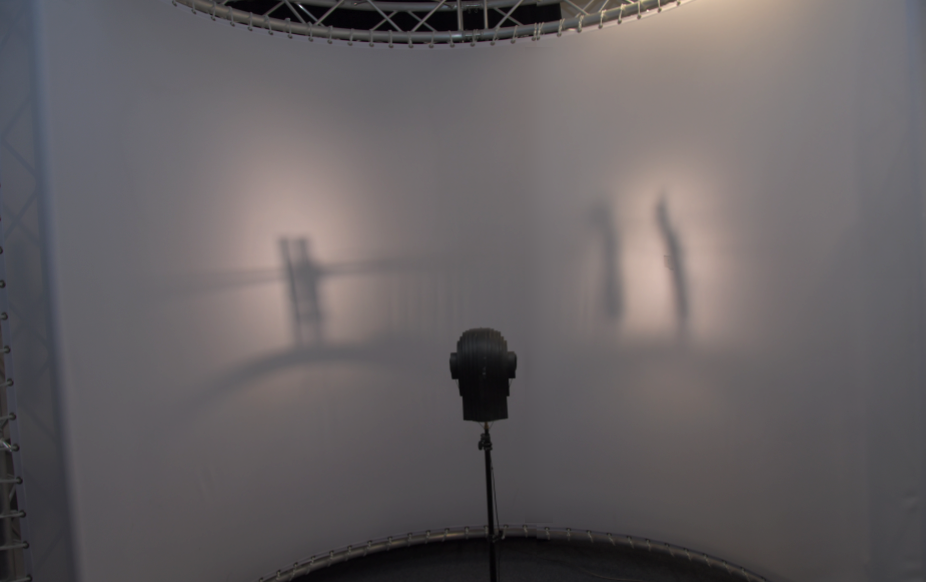}
  \caption{Lab setup of the Soundman head in front of loudspeakers behind a screen. The positions of the speakers are highlighted with external light. See also \cite{Bauer2012SmokeRobotics}.}
  \label{fig:lab_setup}
\end{figure}

To simulate an acoustically close to real-life scenario, we re-recorded the CMU-MOSI corpus in our lab. The experimental setup \cite{Bauer2012SmokeRobotics}, as shown in Fig. \ref{fig:lab_setup}, consists of loudspeakers which are placed around the Soundman wooden head \cite{Davila-Chacon2014ImprovingLocalisation}, behind the white display between 0$^{\circ}$ and 180$^{\circ}$ along the azimuth plane with the same elevation. The Soundman head is 1.6 meters away from the speakers, designing a far-field communication setup. We only use 4 speakers out of 13 speakers. In our previous work \cite{Lakomkin2018EmoRL:Learning, Lakomkin2018OnNetworks} we use the iCub robotic head to test the robustness of our models against the robot's ego noise. However, in this paper, we use the Soundman wooden head to focus on background noise generated by the projectors, computers, air conditioner, power sources as well as noise from airplanes frequently passing nearby, and reverberation noise. The entire recording was done in our lab.

\subsection{Spoken Text Extraction Performance}
\begin{figure}
  \centering
  \includegraphics[width=\linewidth]{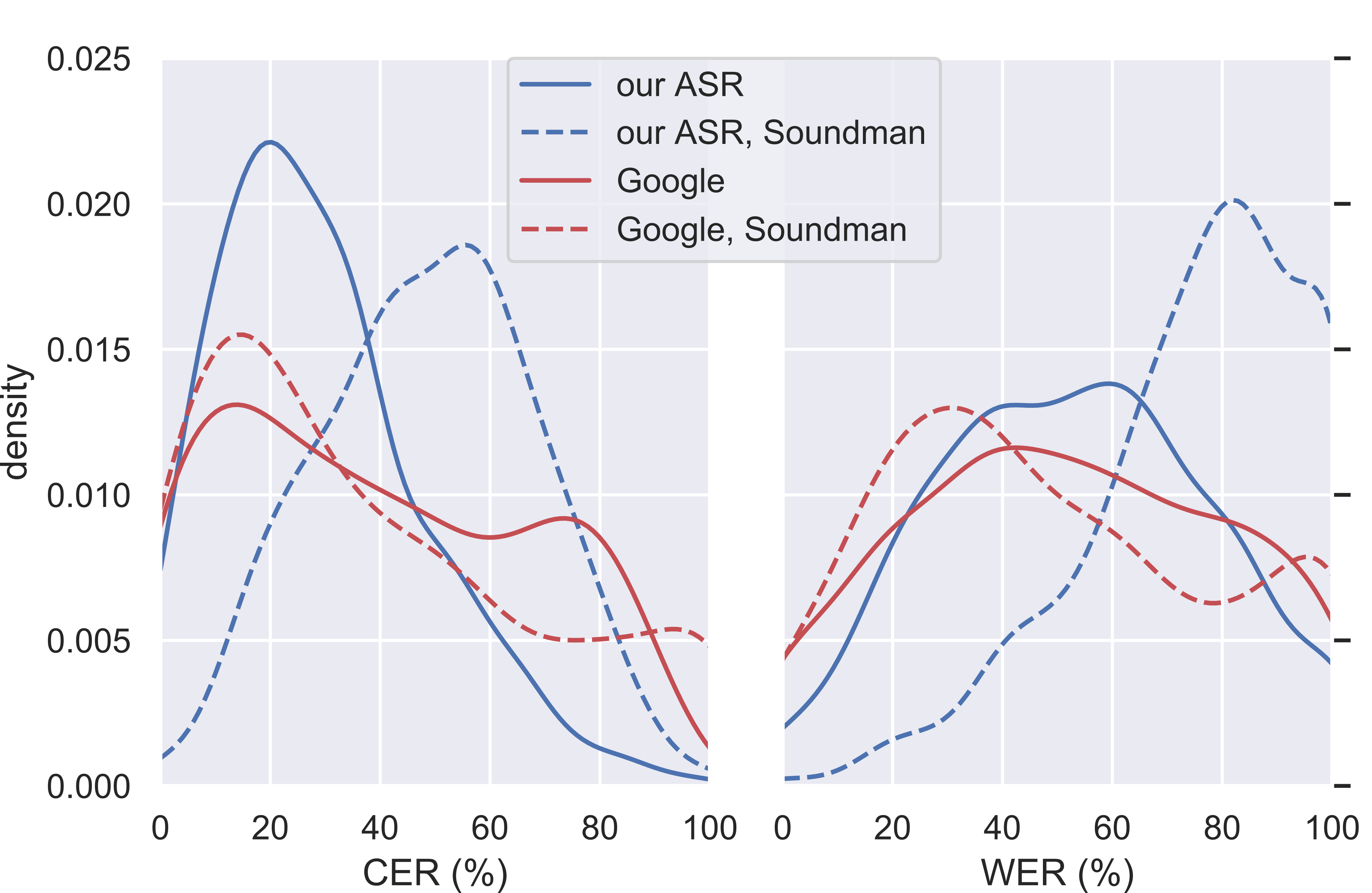}
  \caption{CER and WER density plots for our ASR (blue) and Google ASR (red) evaluated on the MOSI (solid line) and MOSI-Soundman datasets (dashed line).   }
  \label{fig:cer_wer}
\end{figure}
We calculate Word Error Rate (WER) and Character Error Rate (CER) for Google ASR and our ASR model evaluated on the MOSI and MOSI-Soundman datasets. Google ASR has on average 51.3\% WER and 39.7\% CER on the original MOSI dataset, and 49.1\% WER and 38.6\% CER on the MOSI-Soundman. Our ASR model has 53.2\% WER and 28.9\% CER on the MOSI dataset, and 75.7\% WER and 48.6\% CER on the MOSI-Soundman. On the close-field communication (original MOSI) our model shows competitive results, while Google ASR being superior on the far-field scenario.  Figure \ref{fig:cer_wer} shows the density plots of CER for our ASR and Google Web Speech API, demonstrating the difference in the models: the latter system has a lower WER on average and two peaks on the CER density plots, where the right one occurring at the 75\%+ CER area could be explained by language model application. The language model corrects spelling mistakes, but, on the other hand, can change the word completely (for example, names) and increase CER while having better WER.

\subsection{Experimental Results}

\begin{table*}[]
\centering
\caption{Examples from the CMU-MOSI dataset. For each example, we show ground-truth transcription, our ASR model output, sentiment ground-truth and model prediction, and ASR character error rate for this example.}
\vspace{-2mm}
 \label{table:examples}
\begin{tabular}{llccc}
\multicolumn{1}{c}{ground-truth text}                                                                                                                        & \multicolumn{1}{c}{ASR transcription}                                                                                                                       & Sentiment & \begin{tabular}[c]{@{}c@{}}Model\\ output\end{tabular} & CER    \\ \hline
a) I hated it                                                                                                                                                & I haved I                                                                                                                                                   & neg       & pos                                                    & 25\%   \\
\begin{tabular}[c]{@{}l@{}}b) its a pointless scene \\ for the audience and the characters\end{tabular}                                                      & \begin{tabular}[c]{@{}l@{}}it's appointmen seen from the \\ audience tho the characters\end{tabular}                                                        & neg       & pos                                                    & 22.9\% \\ \hline
\begin{tabular}[c]{@{}l@{}}c) anyway oh you can see im still\\  speechless this movie was just beautiful\end{tabular}                                        & \begin{tabular}[c]{@{}l@{}}ay ligt o you can see o stell speechlesses \\ will be was beautiful\end{tabular}                                                 & pos       & pos                                                    & 35.6\% \\
\begin{tabular}[c]{@{}l@{}}d) who have named themselves after the places \\ um to which they have traveled in which \\ i think is a really nice\end{tabular} & \begin{tabular}[c]{@{}l@{}}sen who raee named themselves after the places and to \\ which with travelling which i think is a really nice to or\end{tabular} & pos       & pos                                                    & 32.2\% \\
e) yeah it really is good i mean                                                                                                                             & really is good i mean                                                                                                                                       & pos       & pos                                                    & 26\%   \\ \hline
f) it was terrible                                                                                                                                           & ws terrible i                                                                                                                                               & neg       & neg                                                    & 30.7\% \\
\begin{tabular}[c]{@{}l@{}}g) but if you are a child who grew up in that time \\ period youre not going to enjoy this movie\\  very much\end{tabular}        & \begin{tabular}[c]{@{}l@{}}but if youere a child who grew up in that time cerrod \\ you're not going to enjoy this movvie very much because\end{tabular}    & neg       & neg                                                    & 13.9\% \\ \hline
\begin{tabular}[c]{@{}l@{}}h) but nevertheless another really \\ cool thing about this movie\end{tabular}                                                    & but never was anoter really callring about this ruie                                                                                                        & pos       & neg                                                    & 27.5\% \\
\begin{tabular}[c]{@{}l@{}}i) um that being said you can tell that lot people \\ were having fun with this\end{tabular}                                      & \begin{tabular}[c]{@{}l@{}}that being said you can sell that lot people were \\ having to phumbl ois\end{tabular}                                           & pos       & neg                                                    & 20.3\%
\end{tabular}
\end{table*}
\par
We present the results of our experiments in Table \ref{table:results}. We observe that the character-level mLSTM model pretrained on the Amazon reviews with a simple logistic regression can outperform all the previously published methods, which are more complex and integrate multiple modalities. One argument could be that this mLSTM model even though trained unsupervised has seen significantly more textual data than any work we compared it with. On the other hand, the pretrained word embeddings used in \cite{Zadeh2017TensorAnalysis}, for instance, can be an example of encoded external knowledge. Therefore, we believe that this pre-trained mLSTM model is extremely useful for the spoken sentiment recognition task. In addition, we achieve 69.9\% and 69.6\% accuracy score using the same model, but with the character output of our trained ASR system and Google ASR respectively. Interestingly, the fusion of these two models yields a significant gain in performance, achieving 72.3\% accuracy and 73.6\% accuracy by adding acoustic features. These ASR models make different and uncorrelated mistakes and this can explain the boost of combining them. Overall, the performance gap between the previous best-reported model using acoustic and linguistic modalities and our setup is around 1\% accuracy and 1.3\% F-score. However, in our experiments, we did not use ground-truth transcriptions, but only raw and processed with ASR acoustic signal ones. Our results show that Google ASR is robust to the change of recording conditions and we get similar results on the MOSI and MOSI-Soundman data, while our ASR system performs significantly better on the original MOSI data. We hypothesize that our data augmentation pipeline needs to be improved further by simulating different room conditions during random training to achieve better results in noisy and reverberant conditions for far-field communication. 

\par We provide several examples from the MOSI dataset in table \ref{table:examples}. The examples \textit{b} and \textit{i} demonstrate a situation when the ASR did not recognize correctly the key word, which changes the sentiment of the phrase completely. However, the examples \textit{d} or \textit{e} demonstrate that even with a relative high CER value of 32\% the character-level model can tolerate those errors and correctly classify the sentiment.

\subsection{Importance of ground-truth transcriptions for word-level  sentiment model}

As we observed a significant drop in performance when using spoken text extracted from Google ASR or our own ASR system, we performed an additional experiment using a word-level model with architecture similar to \cite{Kim2014ConvolutionalClassificationb}. It consists of a 1-dimensional convolution network with 100 filters of sizes 2,3,4 and 5, followed by a fully connected layer with 400 units and an output node with a sigmoid activation for binary sentiment modelling. We achieved 74.7\% accuracy on the MOSI test set, similar to spoken text-only results \cite{Zadeh2017TensorAnalysis}. However, if we substitute test set transcriptions with the Google ASR results, we observe a drop to 56.8\% accuracy, which can be a sign of significant overfitting to the specific words and text modality in general. This result is the additional testimony that it is crucial to take into account potential ASR mistakes during training to achieve robust sentiment recognition in practice.

\section{CONCLUSIONS}
\par We addressed spoken sentiment recognition
in conditions when ground-truth text is not available.  Multiple previous works demonstrated that the linguistic features dominate audio-visual input in sentiment and emotion recognition tasks. However, we note that those systems were trained and evaluated using human-transcribed text and, practically, we are not able to use it, for example, during human-robot interaction, when spoken text should be recognized in real time. We demonstrated the discrepancy in performance when ground-truth transcriptions are not present as input and ASR output is used instead on two models: character-level mLSTM with the linear classifier and word-level CNN. However, we observe significant improvements over the acoustic-only baseline on the original and re-recorded MOSI data by adding the ASR hypothesis, which shows that the linguistic modality is still crucial to achieving high-performance sentiment recognition.

\par In future work, we plan to investigate further ways to integrate multiple ASR hypotheses for robust sentiment and emotion recognition. The ensemble of Google Web Speech API and our ASR model show a significant boost in performance indicating the need of having a diverse set of hypotheses to make a better judgement of the affective state of a speaker. Character-level representations learned from the unsupervised language modelling task show very promising performance and we plan to research further whether a similar approach can be transferred to learning robust acoustic representations.
\par The demo, our pre-trained ASR model and its parameters are available at \url{https://github.com/EgorLakomkin/icra_2019_speech}





\section*{ACKNOWLEDGMENT}
The authors thank Erik Strahl for his continuous support with the experimental setup, Julia Lakomkina for her help with illustrations, and Tayfun Alpay for his help in preparing the paper.
This project has received funding from the European Union's Horizon 2020 research and innovation programme under the Marie Sklodowska-Curie grant agreement No 642667 (SECURE) and the German Research Foundation DFG under
project CML (TRR 169).



\bibliography{root}

\end{document}